\documentclass[12pt]{scaleai-paper}
\usepackage[square,sort&compress,numbers]{natbib}

\usepackage{mathpazo}

\usepackage[scaled=0.92]{helvet} 
\usepackage{fontawesome5}
\usepackage{subcaption}
\usepackage{hyperref}
\usepackage{tcolorbox}
\usepackage{graphicx}

\usepackage{hyperref}       
\hypersetup{
    colorlinks=true,
	linkcolor=blue,
	filecolor=magenta,      
	urlcolor=blue,
	citecolor=black,
	pdfinfo={
        Title={Online Rubrics Elicitation from Pairwise Comparisons},
        Subject={rubrics, reinforcement learning, post-training},
        Keywords={post-training, reinforcement learning, scale ai, rubrics, grpo},
    }
}
\usepackage{longtable}
\usepackage{tabularx}
\usepackage[utf8]{inputenc} 
\usepackage[T1]{fontenc}    
\usepackage{hyperref}       
\usepackage{url}            
\usepackage{booktabs}       
\usepackage{amsfonts}       
\usepackage{nicefrac}       
\usepackage{microtype}      
\usepackage{booktabs}
\usepackage{multirow}
\usepackage{amsmath}
\usepackage[ruled,vlined]{algorithm2e}
\usepackage[capitalise]{cleveref}
\usepackage{paralist}
\usepackage{amsthm}
\usepackage{xspace}
\usepackage{graphicx}
\usepackage[table]{xcolor}
\usepackage{array}

\usepackage{enumitem}
\setlist[itemize]{leftmargin=*}
\setlist[enumerate]{leftmargin=*}

\usepackage{graphicx}
\usepackage{tcolorbox}

\usepackage{amsmath}
\usepackage{soul}
\usepackage{cleveref}
\usepackage{listings}
\usepackage{tikz}
\usepackage{eso-pic}
\usepackage{wrapfig}

\let\svthefootnote\thefootnote
\newcommand\freefootnote[1]{%
  \let\thefootnote\relax%
  \footnotetext{#1}%
  \let\thefootnote\svthefootnote%
}
\makeatletter
\renewcommand\AB@affilsepx{, \protect\Affilfont}
\makeatother

\newcommand{\ours}{OnlineRubrics\xspace}
\newcommand{\llmextractor}{$\text{LLM}_\text{extractor}$\xspace}
\newcommand{\llmgrader}{$\text{LLM}_\text{grader}$\xspace}

\title{Online Rubrics Elicitation from Pairwise Comparisons}

\author{MohammadHossein Rezaei$^{1,2,*}$}
\author{Robert Vacareanu$^1$}
\author{Zihao Wang$^1$}
\author{Clinton Wang$^1$}
\author{Bing Liu$^1$}
\author{Yunzhong He$^1$}
\author{and Afra Feyza Akyürek$^1$}

\affil{$^1$Scale AI, $^2$University of Arizona
\\ \vspace{0.5em} $^*$\textit{Work done during internship at Scale AI}
}

\newcommand{\authoremail}{%
  \vspace{-1.5em}
    \faEnvelope\  \texttt{feyza.akyurek@scale.com} \quad 
    \faGlobe\  \url{https://scale.com/research/onlinerubrics}
}

\let\oldcite\cite
\let\cite\citet
\let\citep\oldcite

\newtheorem{proposition}{Proposition}
\crefname{proposition}{Proposition}{Propositions}

\crefname{figure}{Figure}{Figures}
\Crefname{figure}{Figure}{Figures}

\crefname{table}{Table}{Tables}
\Crefname{table}{Table}{Tables}

\crefname{section}{Section}{Sections}
\Crefname{section}{Section}{Sections}

\crefname{equation}{Equation}{Equations}
\Crefname{equation}{Equation}{Equations}
\crefname{appendix}{Appendix}{Appendices}

\begin{document}
\maketitle
\authoremail

\begin{abstract}
Rubrics provide a flexible way to train LLMs on open-ended long-form answers where verifiable rewards are not applicable and human preferences provide coarse signals.
   Prior work shows that reinforcement learning with rubric-based rewards leads to consistent gains in LLM post-training. Most existing approaches rely on rubrics that remain static over the course of training. Such static rubrics, however, are vulnerable to reward-hacking type behaviors and fail to capture emergent desiderata that arise during training.
   We introduce Online Rubrics Elicitation (\ours), a method that dynamically curates evaluation criteria in an \textit{online} manner through pairwise comparisons of responses from current and reference policies. This online process enables continuous identification and mitigation of errors as training proceeds. Empirically, this approach yields consistent improvements of up to 8\% over training exclusively with static rubrics across AlpacaEval, GPQA, ArenaHard as well as the validation sets of expert questions and rubrics. We qualitatively analyze the elicited criteria and identify prominent themes such as transparency, practicality, organization, and reasoning.
\end{abstract}

\section{Introduction}
\label{sec:introduction}

Recent advances in reinforcement learning are reshaping the traditional post-training recipe. \citet{deepseekr1} demonstrated that supervised fine-tuning on instructions can be skipped altogether, with policies (e.g. R1-Zero) trained directly via reinforcement learning, disrupting the way researchers think about post-training. Since then, much of the focus has shifted towards reinforcement learning. However, R1-Zero was trained only using verifiable rewards; the final response is easily gradable, think of a number or code snippet with unit tests, which is only applicable to limited domains.

To accommodate broader settings, rubric-based scoring for reinforcement learning emerges as an alternative way for reward modeling, particularly for long-form responses \citep{viswanathan2025checklistsbetterrewardmodels, rubricsasrewards, huang2025reinforcementlearningrubricanchors, anugraha2025r3}. Rubrics are comprised of a list of input-specific criteria that characterizes an ideal response; one example criterion in the finance domain is \textit{``States shocking basis causes nonlinear effects in margin calls''}. Each criterion has an importance weight: satisfying positively weighted criteria yields reward, while satisfying negatively weighted criteria yields penalty. During training, an LLM-based grader evaluates a response against each criterion in the rubric, producing binary satisfaction scores; and the overall score is the weighted average of these grades. This framework extends reinforcement learning to both verifiable and non-verifiable aspects of responses, spanning generalist and expert domains alike.

Rubrics often emphasize the desired behaviors with less coverage of undesired properties. \textit{Offline rubrics} created a priori, human-written or synthetic, cannot realistically cover every unexpected (and desired) pattern. Fixed checklists \citep{wang-etal-2024-interpretable} to enforce generally helpful patterns e.g. truthfulness, instruction following or relevance, fall short in preventing nuanced errors. For example, \cite{huang2025reinforcementlearningrubricanchors} identifies ``self-praising'' as one emerging pattern during reinforcement learning from rubrics, think of including \textit{``The following advice is the most relevant''} as part of the response; these praises often fool the LLM-based grader into believing that the given response is indeed relevant. Such patterns are especially difficult for generic ``catch-all'' rubrics to reveal when they are sample-specific. Moreover, correct traits in some generations can go unnoticed if not readily rewarded by the existing offline rubrics.

\begin{figure}[t]
    \centering
    \includegraphics[width=0.97\textwidth]{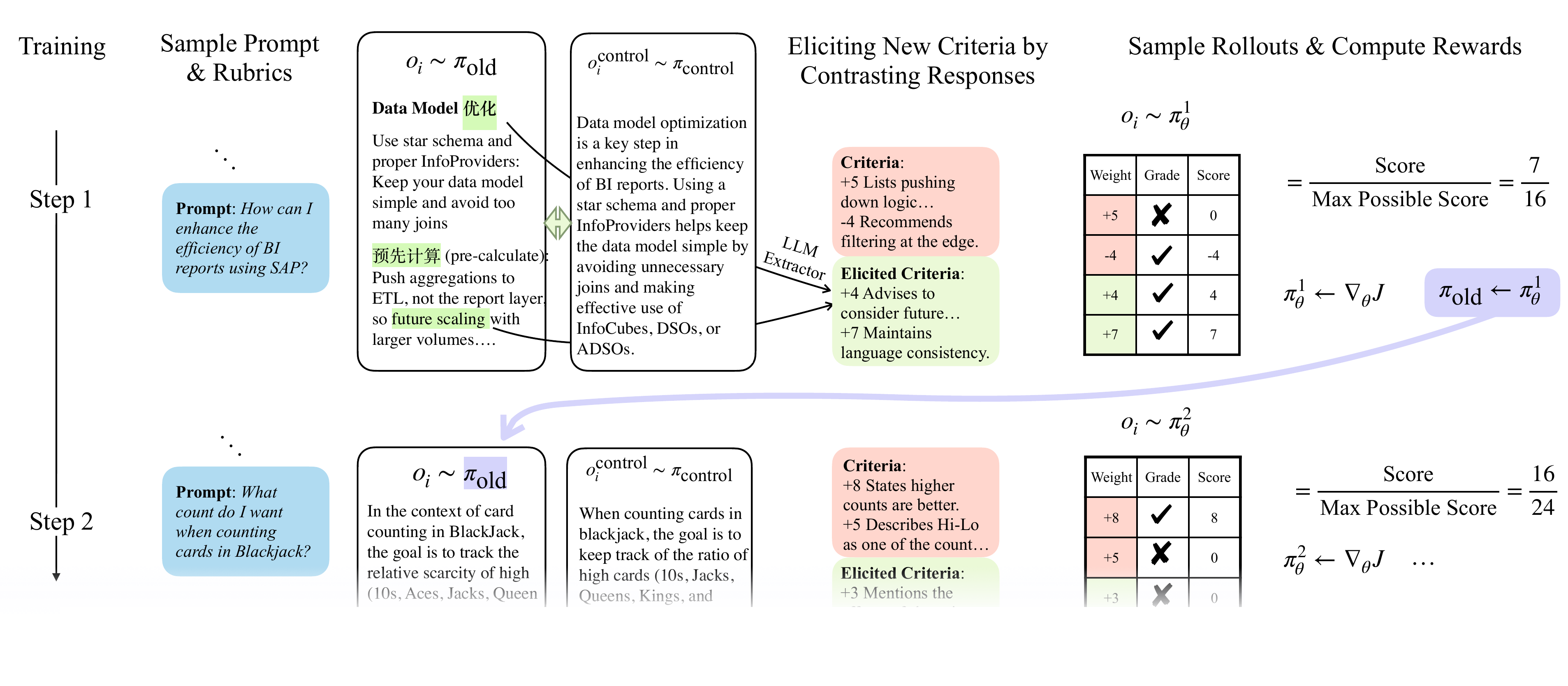} 
    \caption{At any step during training, \ours starts off by considering a pair of responses, one of which is from the current policy before updates and another from a \textit{control} model e.g. reference model. We follow with LLM-based rubrics elicitation and deduplication steps to generate a set of \textit{elicited criteria}. These criteria along with existing criteria (e.g. human-written or synthetic) are used to create the reward in the policy gradient algorithm.}
    \label{fig:method}
\end{figure}

We introduce \ours, a framework for eliciting evaluation criteria dynamically via pairwise comparisons. \ours leverages a pair of responses in creating additional criteria where the responses are sampled from the current policy and a control model. Our work, as depicted in \cref{fig:method}, is inspired by the large body of literature on preference learning \citep{akrour2011preference, furnkranz2012preference, pmlr-v32-schoenauer14} and pairwise reward modeling \citep{christiano2017deep,stiennon2020learning,ouyang2022training}. While LLMs are imperfect judges of quality \citep{gu2024survey}, we found that pairwise comparisons are easier to make for the models when identifying new criteria than directly making a quality assessment or creating new criteria by considering a single response (point-wise elicitation). The additional criteria simply augments the existing rubric, enabling seamless integration of \ours with any rubric-based scoring mechanism. 

In training and evaluating our approach, we curate two datasets for expert (scientific use-cases) and generalist domains. We additionally conduct out-of-distribution evaluations using public benchmarks, comparing different approaches to reward estimation. \ours results in absolute gains of up to 25\% over the initial instruct model across various benchmarks including GPQA-Diamond \citep{rein2024gpqa}, GSM8K \cite{gsm8k}, AlpacaEval \cite{alpaca_eval}, and Arena-Hard \citep{arenahard2024}.

\section{Related Work}
\label{sec:relatedwork}

\paragraph{Reward Modeling} The dominant paradigm in LLM alignment is to learn a reward function from feedback.
Foundational work in Reinforcement Learning from Human Feedback (RLHF) established the use of pairwise preference comparisons--preferred over less robust pointwise scores--to train an explicit reward model \citep{ouyang2022training, stiennon2020learning}. This process was later simplified by methods like Direct Preference Optimization (DPO;~\cite{dpo}), which bypasses the explicit reward model and optimizes policies directly on preference data. Methods for generating feedback have also advanced: \cite{cai}, for example, pioneered the use of AI feedback (RLAIF) by leveraging a fixed set of principles for model self-feedback. More recently, research has focused on improving the reward model's intrinsic capability. \cite{deepseekgrm} established inference-time scaling laws for generalist reward models, boosting performance with added computation, while \cite{j1} incentivizes faithful evaluation by training LLM judges to generate reasoning.

While preference-based rewards provide flexible but often fuzzy signals, verifiable rewards offer exact supervision whenever the outcome can be automatically checked. Reinforcement Learning with Verifiable Rewards (RLVR) improves reasoning by optimizing policies against automatically checkable outcomes, such as numeric answers or unit-tested code. Recent work has shown its effectiveness across various domains: DeepSeek-R1 \citep{deepseekr1} and General-Reasoner \citep{general_reasoner} achieved strong results on benchmarks such as GSM8K \citep{gsm8k}, MMLU \citep{hendrycks2021measuring}, and GPQA \citep{rein2024gpqa}. In medicine, \cite{med-rlvr} enabled a 3B model to reach expert-level performance. Foundational studies confirm that RLVR incentivizes correct reasoning processes, not just correct answers \cite{wen2025}. Despite these strengths, RLVR does not extend to open-ended domains where correctness cannot be automatically verified.

\paragraph{Multi-Objective Alignment}
Beyond single-reward formulations, recent research has explored \emph{multi-objective RLHF} approaches that optimize across several criteria simultaneously. Safe RLHF \citep{dai2023safe} decouples helpfulness and harmlessness rewards and balances them using constrained optimization. Gradient-Adaptive Policy Optimization (GAPO) \citep{li2025gapo} employs multiple-gradient descent to achieve Pareto-optimal trade-offs across competing objectives, while \cite{lu2025learning} proposes dynamically adjusting reward weights online. Similarly, conditional reward modeling \citep{cai2024internlm2} allows a single reward model to flexibly apply different principles depending on context in training their evaluator LLM. These works highlight growing recognition that LLM alignment requires balancing diverse objectives which is closely related to our focus on dynamically eliciting new rubrics.

\paragraph{Evaluating and Training with Rubrics} 
Recent work has extended the concept of verifiable rewards from domains like math and coding to more open-ended tasks by using rubrics for structured evaluation. This rubric-based approach has been adopted in various benchmarks for both expert \citep{healthbench, paperbench} and generalist domains \citep{multichallenge}. Beyond evaluation, rubrics are now increasingly used as direct reward signals for reinforcement learning. Using structured rubrics as a direct reward has proven effective in both expert reasoning \citep{rubricsasrewards} and generalist alignment \citep{viswanathan2025checklistsbetterrewardmodels}. A diverse set of rubrics has also been used to train a single, robust reward model that generalizes across various domains \citep{anugraha2025r3}. Our work complements these methods; instead of using a static rubric or training a rubric-agnostic model, \ours dynamically augments criteria online to adapt to the policy's emergent behaviors.

\section{Background}
\label{sec:background}
Rubrics are often used as drop-in replacement for rewards in any policy gradient learning algorithm. 

\subsection{Training Setup}

In this work, we used the GRPO algorithm \citep{shao2024deepseekmath} maximizing the following objective 
{\small
\begin{equation}
\mathcal{L}_{\text{GRPO}}(\theta) \;=\;
\mathbb{E}_{i \sim \mathcal{D},\, j \sim \mathcal{G}_i}\Bigg[
\min \Big(
r_{i,j}(\theta) \, \hat{A}_{i,j}^{\text{group}}, \;
\text{clip}\big(r_{i,j}(\theta),\, 1-\epsilon,\, 1+\epsilon\big) \, \hat{A}_{i,j}^{\text{group}}
\Big) - \beta \mathbb{D}_{KL}\Big(\pi_\theta ||\pi_{ref} \Big)
\Bigg]
\label{eq:grpo_objective}
\end{equation}
}

where $r_{i,j}(\theta) = 
\frac{\pi_\theta(o_{i,j}\mid x_i)}{\pi_{\theta_{\text{old}}}(o_{i,j}\mid x_i)}$ is the probability ratio, and advantages are calculated as normalized rewards:

\begin{equation}
    \hat{A}_{i,j}^{\text{group}} = \frac{R_j - \text{mean}(\textbf{R})}{\text{std}(\textbf{R})}
    \label{eq:advantages}
\end{equation}

where $\mathcal{D} = \{x_i, \mathcal{C}_i\}$ is the set of training prompts and criteria, $j$ indexes the output samples $o_j$ from the group $o_j \sim \mathcal{G}_i$, $\pi_{\theta_{\text{old}}}$ is the policy before the update, $\pi_{\theta}$ the target policy. The rewards are computed independently for each $o_j$ in the group and denoted by $\textbf{R} = \{R_1, R_2, \dots, R_G\}$ where $G$ is the group size.

In this work, we will assume that the true reward $U$ can be modeled as a function of latent criteria and argue in \cref{sec:formalmotivation} that for optimal modeling of the true reward all criteria should be elicited.

\subsection{Rubric Based Rewards}
In RLHF, reward signals in LLM training are traditionally modeled after human preferences with an explicit reward model in PPO \citep{schulman2017proximal} and GRPO or implicitly in DPO. In the case of queries where quick verification of the final answer is possible (i.e. numeric or short answer), exact match replaces human preferences for reward. More recently, rubrics for evaluating long-form answers are being used for calculating final scores \citep{rubricsasrewards, huang2025reinforcementlearningrubricanchors, viswanathan2025checklistsbetterrewardmodels} where an LLM-based grader (denoted by \llmgrader) evaluates a response against each criteria to compute $R_{j}$ in \cref{eq:rubric_reward}:

\begin{equation}
    R_j = q\Big(\text{LLM}_\text{grader}\Big(o_j, x_i, \mathcal{C}_i\Big)\Big)
    \label{eq:rubric_reward}
\end{equation}

where $\mathcal{C}_i = \{(c_1,w_1), (c_2, w_2), \dots, (c_d, w_d)\}$ is a collection of criteria with corresponding importance weights that describe an ideal response to the prompt, and $q$ is an reduction function. The judge \llmgrader \citep{NEURIPS2023_91f18a12} evaluates the output $o_j$ against each criterion in $C_i$ and produces a list of binary outcomes which are then reduced to a single scalar value by $q$ using the weights, if applicable. In this work we implement the reduction function as a weighted sum of the grades normalized by the total possible maximum score:
\begin{equation}
    q(x,o,\mathcal{C}) = \frac{w^\top \text{LLM}_\text{grader}(x,o,\mathcal{C})}{\sum_{k:w_k>0} w_k} 
    \label{eq:rubric_reward_reduction}
\end{equation}

where \llmgrader$(x,o,\mathcal{C}) \in \{0,1\}^d$ is the binary grades corresponding to each criterion.

\section{Online Rubric Elicitation}
\label{sec:pairwise}

\begin{algorithm}[t]
\caption{Online Rubric Eliciting (\ours)}
\KwIn{Policy $\pi_\theta$, control policy $\pi_{\text{control}}$, dataset $\mathcal{D}$, extraction prompt $P_e$, hyperparameter $M$}
\For{$step = 1, 2, \dots, N$}{
    Sample prompts and criteria $\{x_i, \mathcal{C}_i\}$ from $\mathcal{D}$\;
    Update $\pi_\text{old} \gets \pi_\theta$\;
    Generate $M$ candidate responses $\{o_{i,j}\}$ using $\pi_\text{old}$\;
    Generate $M$ candidate responses $\{o^\text{control}_{i,j}\}$ using $\pi_\text{control}$\;
    Initialize $C_i^e \gets \emptyset$\;
    \For{$k = 1, 2, \dots, M$}{
    Extract new criteria $C_{i,k}^e \sim  \text{LLM}_\text{extract}(x_i, o_{i,k}, o^\text{control}_{i,k};P_e)$\;
    $C_i^e \gets C_i^e \cup C_{i,k}^e$\;
    }
    De-duplicate $C_i^e$\;
    Compute rewards using \cref{eq:rubric_reward} and $\mathcal{C} = \mathcal{C}_i \bigcup {C_i^e}$\;
    Compute group advantages $\hat{A}_{i,j}$ \cref{eq:advantages}\;
    Update $\theta$ via policy gradient by maximizing \cref{eq:grpo_objective}
}
\label{algo:ours}
\end{algorithm}

Rubric-based reward calculation provides richer feedback than reward-model-based post-training, yet it fails to mitigate the problems that might emerge during policy gradient updates. Specifically, we observe that initial rubrics tend to represent the desired qualities of an ideal response while putting less emphasis on describing undesired qualities. For example, when the prompt is \textit{``How can I test for the presence of carbon dioxide in a reaction?''} and the rubric is \textit{(+9, The response mentions limewater turning milky)}, both responses \textit{``Bubble the gas through limewater; it turns milky due to calcium carbonate formation. This reaction is specific to CO2''} and \textit{``Bubble the gas through limewater; it turns milky due to calcium carbonate formation, which is slightly soluble in acidic conditions''} receive the full score, while the latter includes technically accurate but unnecessary information unrelated to the prompt. Such mishaps may only be detected as they arise during rollouts. Moreover, emerging desirable qualities (e.g., \textit{``This reaction is specific to CO2''}) that are not currently rewarded by the existing rubric set will be overlooked by the algorithm.

We propose a novel method called \ours that leverages pairwise comparison of candidate responses to derive novel criteria---\ours is designed to capture potential errors and identify useful features. The approach simply augments the set of offline criteria i.e. the portion of the rubric that is created a priori for the specific prompt, with more criteria derived during the training. Our approach is different from recent work that uses a fixed set of criteria (or checklists) \citep{anugraha2025r3} for multiple data points or other procedures to extract rubrics in a pointwise manner by simply considering a prompt \citep{huang2025reinforcementlearningrubricanchors}. \ours drives insights from the pairwise reward modeling literature \citep{bradley1952rank, stiennon2020learning, ouyang2022training}.

\subsection{LLM-based Criteria Elicitation}

\begin{figure}[t]
    \centering
    \includegraphics[width=0.95\textwidth]{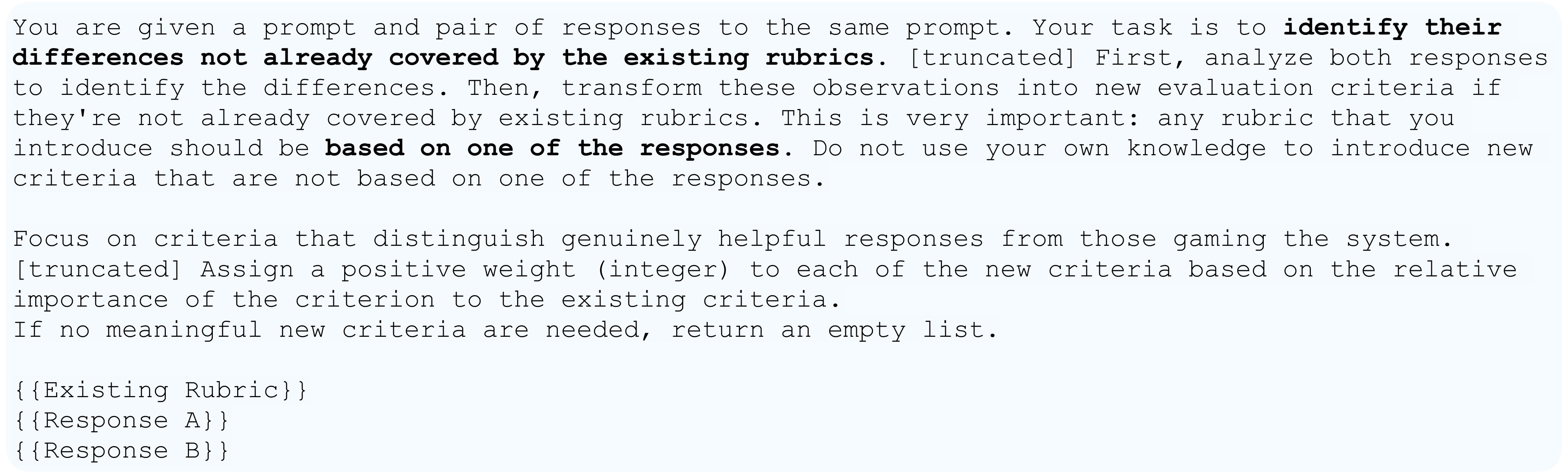}
    \caption{Abbreviated system prompt template used for eliciting new criteria from pairwise response comparisons, see full prompt in \cref{fig:extractor_prompt_full}.}
    \label{fig:extractor_prompt}
\end{figure}

\ours begins with an initial set of \textit{offline criteria} $\mathcal{C}_i$ that may be provided by human annotators or created synthetically. During policy training, at step $t$ before any updates, given a prompt $x_i$ we sample a set of candidate responses from a \textit{control} policy (e.g. the initial policy, $\pi_\text{ref}$, or the policy from the previous step $\pi_\text{old}$) and the current policy $\pi^t_\theta$. We define an LLM-based rubric extractor \llmextractor conditioned on the system prompt $P_e$ (see Figure~\ref{fig:extractor_prompt}) whose task is to identify the differences between a pair of responses $(o_{i,j}, o_{i,j}^\text{control})$ sampled from the current and control policies, respectively, and turn them into useful criteria and corresponding weights. We repeat this procedure independently for each prompt in the batch and augment their corresponding rubrics with the new criteria before the policy parameter update. We provide the procedure in \cref{algo:ours}.

We adopt a two-step approach for criteria elicitation; in the first step, we ask \llmextractor to enumerate the meaningful differences between a pair of responses with references to where these differences arise in the responses. In the second stage, we reduce the criteria that are duplicates or overlap significantly to avoid redundancy following our desiderata in \cref{sec:data}. The system prompt template used to extract rubrics is given in Figure~\ref{fig:extractor_prompt} and the deduplication prompt is available in Figure~\ref{fig:deduplication_prompt}. By default, we compare eight pairs of rollouts from each of the control and current policies and extract about eight criteria at the end of the procedure. 

\paragraph{\ours Variants} We experiment with two variants depending on the source of alternative responses $\pi_\text{control}$ among $\pi_\text{ref}$ or $\pi_\text{old}$. 
We empirically observe in \cref{tab:if-results} that sampling the control set of responses from the $\pi_\text{old}$ also performs quite strongly compared to the setting $\pi_\text{control} = \pi_\text{ref}$ if not better.

\subsection{A Formal Motivation for \ours}
\label{sec:formalmotivation}
Let $f$ be the grades from \llmgrader for the prompt, response and criteria triplet $(x,o,\mathcal{C})$ such that $f(x,o,\mathcal{C}) \in \{0,1\}^{d}$ where $\mathcal{C}$ and $w$ are the set of criteria and weights and $d$ is the size of the criteria. Let also $\mathcal{C}^E$ (explicit) and $\mathcal{C}^I$ (implicit) to denote to the set of criteria in the rubric and those not in the rubric, respectively, and $f_E(x,o)$ to indicate the binary grades for the output $o$ under criteria $\mathcal{C}^E$.
\begin{proposition}
 Suppose that 
 \begin{compactitem}
     \item $\mathcal{C}^*$ is the set of true criteria. $f_*$ can be split into $f_* = (f_E,f_I)$ and $\mathcal{C}^* = \big(\mathcal{C}^E, \mathcal{C}^I\big)$.
     \item The true reward is $U(x,o) = w_E^\top f_E(x,o,\mathcal{C}^E) + w_I^\top f_I(x,o,\mathcal{C}^I)$ and the estimated reward $R_t(x,o) = w_E^\top f_E(x,o)$ at step $t$.
     \item Assuming GRPO style updates, the gradient under the true reward then would be $g_{U} = \mathbb{E}[\nabla_\theta \log \pi_\theta(o|x) U(x,o)]$ and the estimated gradient $g_{R_t} = \mathbb{E}[\nabla_\theta \log \pi_\theta(o|x) R_t(x,o)]$
 \end{compactitem} Then,
\[
 \lVert g_U - g_{R_t} \rVert_2 \leq \sqrt{\mathbb{E}\Big[\big\lVert \nabla_\theta\log_{\pi_\theta} \big\rVert^2\Big]}\lVert w_I \rVert_1
 \]
 \label{prop:gradient}
\end{proposition}

\cref{prop:gradient} shows that the difference between the gradient steps is upper-bounded by $\lVert w_I \rVert_1$ times the expected squared norm of the policy score function. Augmenting the rubric to better approximate the true criterion set leads to better estimation of the true gradient hence improved stability and sample efficiency during training. That said, \ours should be viewed as a step toward tightening the upper bound on the implicit, unmodeled mass $\lVert w_I \rVert_1$, rather than a complete recovery of the true criteria set. Proof is given in Appendix~\ref{app:proof_gradient}.

\section{Datasets}
\label{sec:data}

We trained \ours with two collected rubric datasets: Generalist Rubrics and Expert Rubrics. 
Generalist Rubrics consists of real-world, single-turn prompts contributed with user consent and curated to be safe, rubric-eligible, and generalist in scope. 
For each prompt, human annotators  
authored a prompt-specific rubric composed of weighted, binary-checkable criteria. 

\begin{figure}
    \centering
    \includegraphics[width=0.49\textwidth]{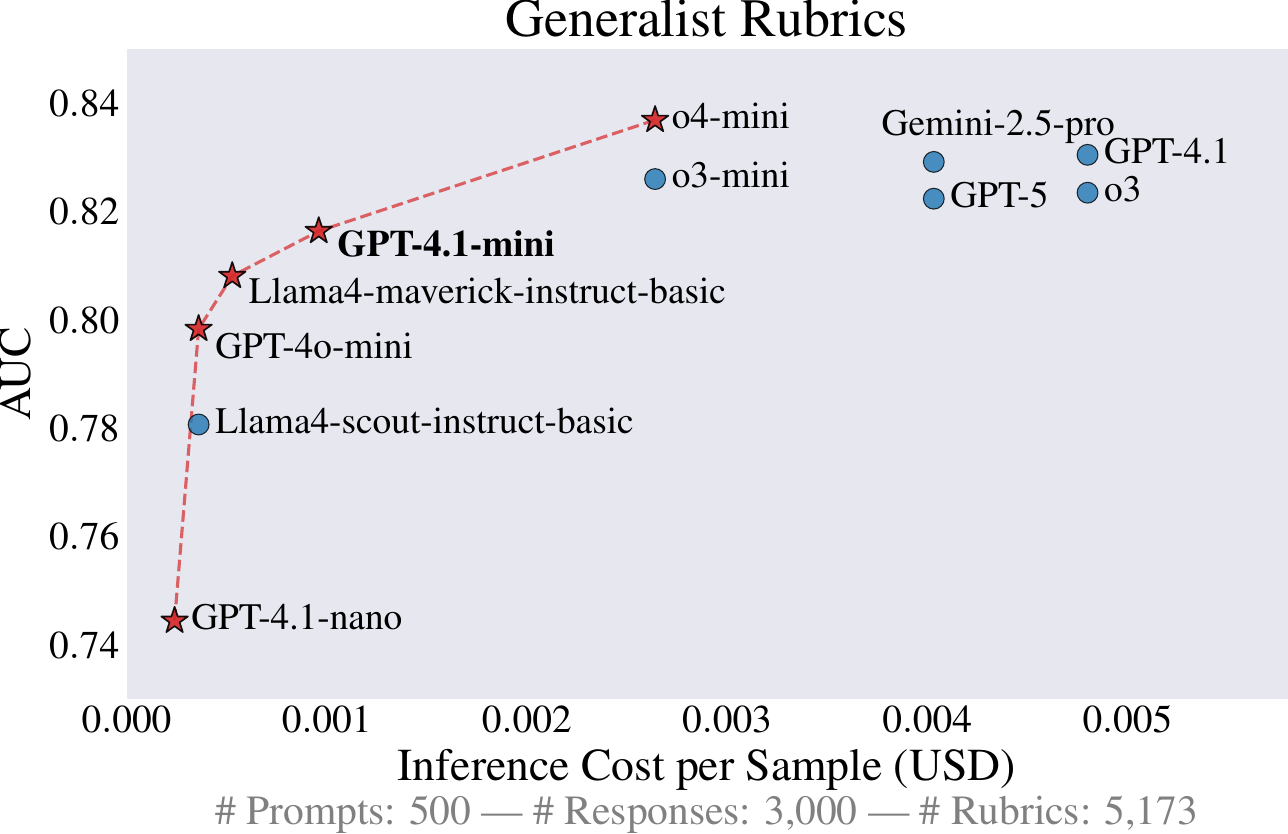}
    \includegraphics[width=0.49\textwidth]{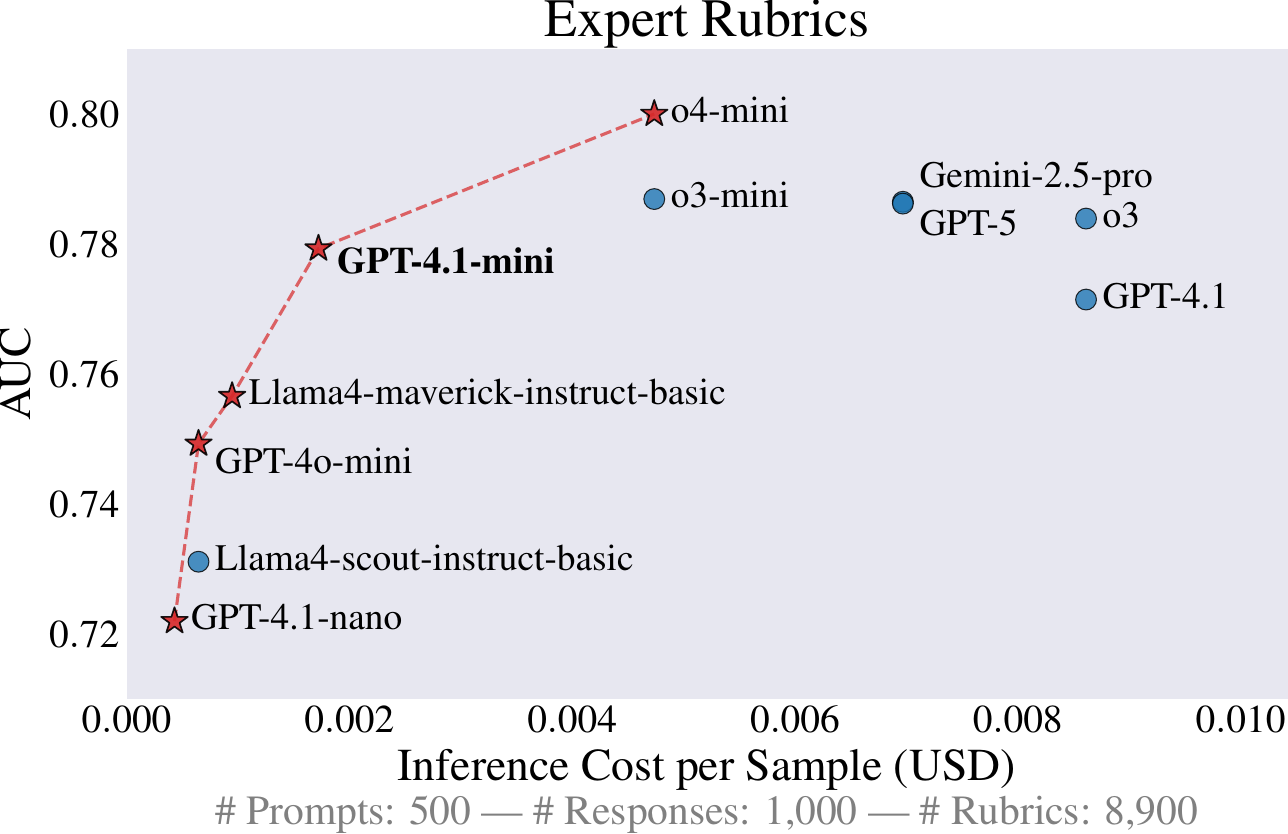}
    \caption{
        Performance of different LLM graders. AUC score is calculated using the receiver operating characteristic (ROC) curve.
        The best grader is the one with the highest AUC score and the lowest inference cost per sample.
        Models on the Pareto frontier (shown as a red dotted line) are the best trade-off between the two metrics. 
        We choose GPT-4.1-mini as our default grader, balancing alignment quality with inference cost.
    }
    \label{fig:verifier-selection}
\end{figure}

\begin{wraptable}{r}{0.49\textwidth}
    \centering
    \caption{Generalist and Expert Rubrics datasets statistics.}
    \label{tab:dataset-stats}
    \small
\begin{tabular}{l r@{\ \ }r r@{} r@{\ \ }r}
\toprule
& \multicolumn{2}{c}{\textbf{Train}} && \multicolumn{2}{c}{\textbf{Eval.}} \\ \cmidrule{2-3} \cmidrule{5-6}
& \# Sam. & \# Rub. && \# Sam. & \# Rub. \\
\midrule
Generalist & 1,500 & 15,528 && 487 & 5,003 \\
Expert & 1,864 & 33,554 && 332 & 5,938 \\
~~~~Math & 584 & 9,512 && 104 & 1,688 \\
~~~~Biology & 506 & 9,863 && 90 & 1,750 \\
~~~~Physics & 314 & 5,631 && 56 & 1,001 \\
~~~~Chemistry & 460 & 8,548 && 82 & 1,499 \\
\bottomrule
\end{tabular}
\end{wraptable}

Expert Rubrics extends the same rubric framework to expert-authored problem sets across Physics, Chemistry, Biology, and Math. 
Each task bundles a prompt, an expert grading rubric with binary-evaluable and weighted criteria, sample model responses, and detailed rubric ratings.

We use a subset of both datasets as evaluation sets and exclude from training. 
Table~\ref{tab:dataset-stats} shows the statistics of the datasets. On average, Generalist set contains 10.4 rubrics per sample and Expert set contains 18.0 rubrics per sample.

Across both datasets, rubrics are human-written and follow the same annotation principles: criteria are \textit{Mutually Exclusive \& Collectively Exhaustive, Atomic, Objective, and Self-Contained}; ensuring they can be verified reliably and used as dense reward signals in offline and online training. See Appendix~\ref{app:data_samples} for data samples.

We evaluate \ours on (1) evaluation sets of both datasets by calculating rubrics score and win rate using Gemini 2.5 Pro~\citep{comanici2025gemini} as an LLM-Judge, and (2) on the following public benchmarks:
GPQA-Diamond \citep{rein2024gpqa}, 
GSM8K \citep{gsm8k}, 
AlpacaEval \citep{alpaca_eval, dubois2024length},
and Arena-Hard \citep{li2024crowdsourced, arenahard2024}. 

\section{Experiments and Results}
\label{sec:results}

We begin by identifying the most effective LLM-based grader for rubric grading in \cref{sec:verifier-selection}. 
Next, we introduce our baselines in \cref{sec:baselines} and report the main results with \ours in \cref{sec:results}. 
Finally, we perform a qualitative analysis of the elicited rubrics in \cref{sec:qualitative-analysis}.

We train Qwen-2.5-7B-Instruct~\citep{qwen2025qwen25technicalreport} with GRPO as the training algorithm on the training data from both Generalist and Expert Rubrics datasets for 3 epochs and evaluate on the eval set of the respective datasets 10 times during each epoch. 
We use \textit{o3-mini} as the \llmextractor and set the number of pairwise comparisons to 8.
Appendix~\ref{sec:experimental_settings} provides the detailed experimental settings.

\subsection{Verifier Selection}
\label{sec:verifier-selection}

Rubrics training requires an LLM grader to evaluate whether an output $o_j$ meets the criteria specified in the rubrics $\mathcal{C}_i$. 
The input to the grader is a (prompt $x_i$, output $o_j$, rubrics $\mathcal{C}_i$) triplet, 
and the output is a sequence of binary scores indicating whether each criterion $c_k \in \mathcal{C}_i$ is satisfied by the output.
Although grading is assumed to be easier than generation~\citep{stechly2024selfverificationlimitationslargelanguage},
it is still a challenging task for LLMs and remains under-explored in previous work on rubrics due to the lack of human-annotated data with fine-grained rubric-level scores. However, different LLM graders have different evaluation capabilities, 
which can significantly affect the training of rubrics-based models.
To address this, we have collected human evaluations of the original human-written rubrics for 2-6 sampled responses per prompt for 500 prompts for each of Expert and Generalist sets.

Using this dataset, we evaluate the performance of several LLM graders and present the results in Figure~\ref{fig:verifier-selection}.
Given that during the rubrics-based training, 
we need to evaluate multiple rollouts for each prompt,
it is important to choose a grader with a low inference cost per sample.
We calculate the inference cost per sample by dividing the total inference cost by the total number of samples.

Perhaps unsurprisingly, we find that all verifiers perform better on the Generalist dataset than the Expert dataset (average AUC score of 0.811 vs 0.768). 
Interestingly, the Pareto frontier for the Generalist dataset is the same as the Pareto frontier for the Expert dataset.
This suggests that the relative performance of the verifiers is not affected by the domain. We choose GPT-4.1-mini as our default grader, balancing the alignment with grades with inference costs.

\begin{figure}
    \centering
    \includegraphics[width=0.49\textwidth]{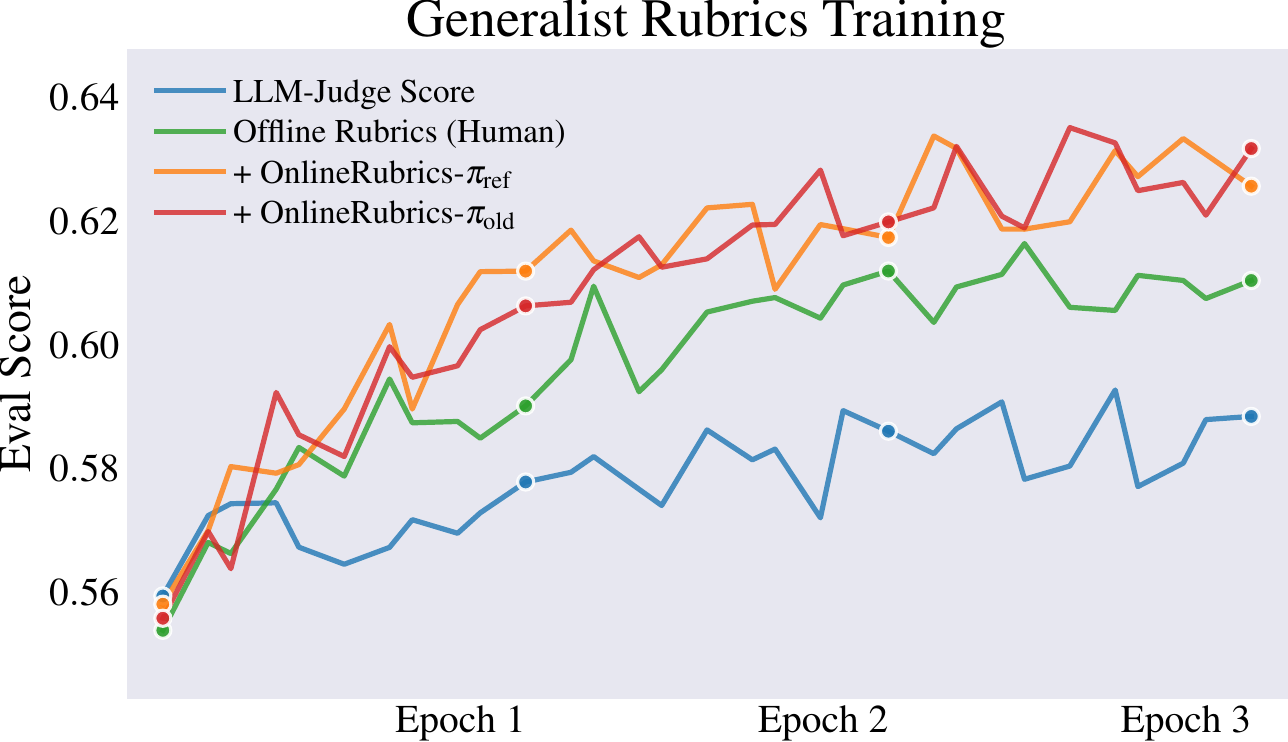}
    \includegraphics[width=0.49\textwidth]{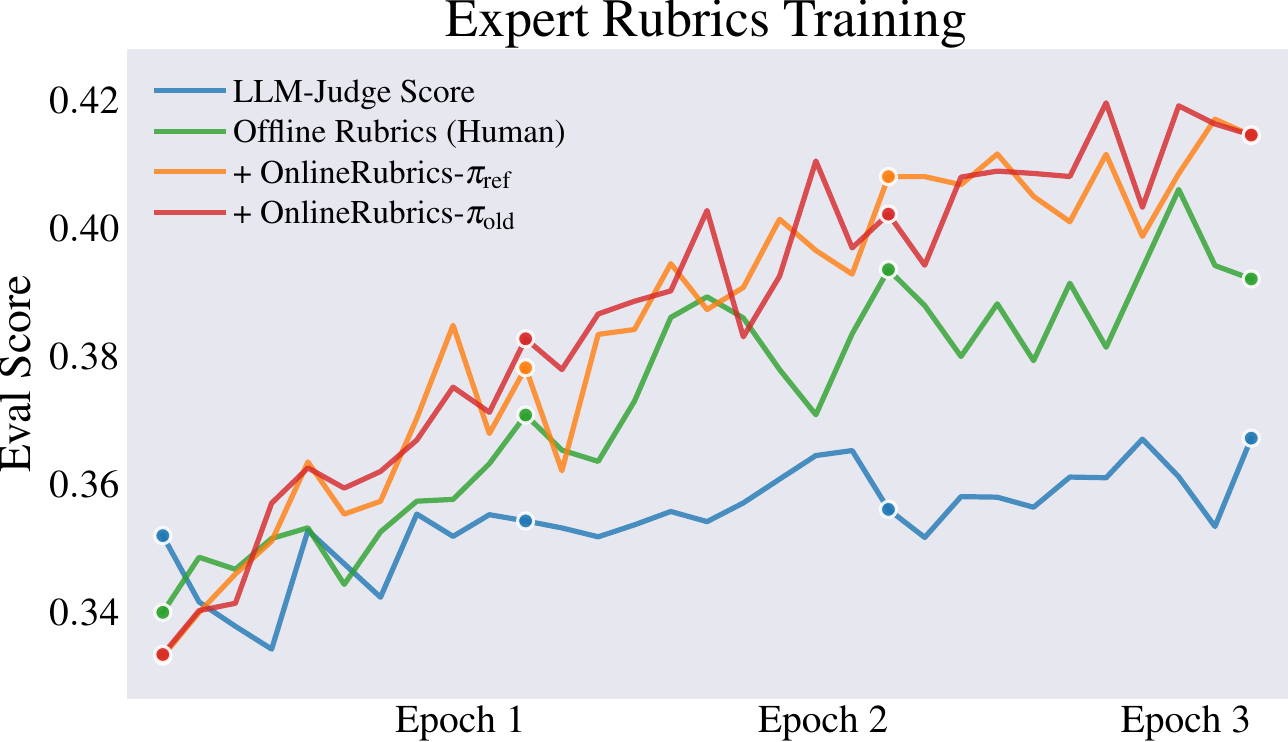}

    \caption{
        Results on the evaluation set of the Generalist and Expert datasets during training (higher is better).
        The evaluation set is fixed and does not contain any elicited rubrics.
        Both \ours methods outperform using Offline Rubrics (Human) or LLM-judge Score (a Likert scale).
    }
    \label{fig:training-evaluation-curves}
\end{figure}

\subsection{Baselines}
\label{sec:baselines}

We compare our methods with the following baselines:

\textbf{LLM-Judge Score} 
We train the model by only using an LLM-judge to grade the responses on a Likert scale without any rubrics.
The input to the LLM-judge is a prompt-response pair $(x_i, o_{j})$, and the output is a Likert score that is converted to a reward $R_{i,j}$ using a linear mapping.
We experiment with o3-mini as the LLM-judge. The prompt is given in Appendix~\ref{app:system_prompts}.

\textbf{Offline Rubrics (Synthetic)}
We use the same prompts available in the Generalist and Expert Rubrics datasets.
However, instead of using human-written rubrics, we synthetically create rubrics using o3-mini.
See the prompt in \cref{app:system_prompts}.

\textbf{Offline Rubrics (Human)} 
We train the model with human-written rubrics from the Generalist and Professional Rubrics datasets.
As we shall see, using human-written rubrics, often significantly, is better than using synthetic rubrics across the benchmarks we evaluate.

\textbf{Universal Requirements} 
As discussed in \cref{sec:relatedwork},
previous work argued that adding a fixed set of criteria to all samples helps the model to make training more stable and prevent reward hacking. We use the same universal requirements as in \cite{viswanathan2025checklistsbetterrewardmodels} and show \ours, which elicits sample-grounded rubrics online, outperforms these universal requirements.

\textbf{Point-wise Elicitation} 
In order to show the effectiveness of pairwise comparison, we also extract rubrics point-wise using the same extractor model. 
The input to the extractor is prompt $x_i$, a response $o_j$ from the reference policy, and existing rubrics $\mathcal{C}_i$.
The output is a set of criteria $C_i^e$ that we add to the human-written rubrics $\mathcal{C}_i$.

\begin{table*}
    \centering
    \begin{tabular}{l rr r@{} rr r@{} c}
\toprule
 & \multicolumn{2}{c}{\textbf{Generalist Rub.}} && \multicolumn{2}{c}{\textbf{Alpaca-Eval}} && \multicolumn{1}{c}{\textbf{Arena-Hard}} \\ \cmidrule{2-3} \cmidrule{5-6} \cmidrule{8-8}
\textbf{Model} & \multicolumn{1}{c}{Score} & \multicolumn{1}{c}{WR} && \multicolumn{1}{c}{WR} & \multicolumn{1}{c}{LC-WR} && \multicolumn{1}{c}{WR} \\
\rowcolor{gray!20} \multicolumn{8}{c}{\textit{Baselines}} \\
Qwen-2.5-7B-Instruct & 55.4 & 39.0 && 30.0 & 28.2 && 50.0 \\
~~~~+ LLM-Judge Score & 58.8 & 51.3 && 42.2 & 26.9 && 51.0 \\
~~~~+ Offline Rubrics (Synthetic) & 58.8 & 52.8 && 39.5 & 28.2 && 51.5 \\
~~~~+ Offline Rubrics (Human-written) & 61.0 & 62.2 && 46.4 & 28.0 && 52.4 \\
~~~~~~~~+ Universal Requirements & 59.4 & 59.1 && 44.4 & 30.3 && 53.8 \\
~~~~~~~~+ Pointwise Extraction & \underline{62.9} & 64.9 && 48.1 & 29.4 && 51.1 \\ 
\rowcolor{gray!20} \multicolumn{8}{c}{\textit{Our Methods}} \\
~~~~~~~~+ \ours-$\pi_\text{ref}$ & 62.7 & \underline{67.6} && \underline{54.0} & \textbf{31.5} && \underline{55.7} \\ 
~~~~~~~~+ \ours-$\pi_\text{old}$ & \textbf{63.2} & \textbf{68.2} && \textbf{55.0} & \underline{30.4} && \textbf{56.5} \\
\bottomrule

\end{tabular}
    \caption{Results on the instruction-following benchmarks.
    WR stands for Win Rate and LC-WR is Length-Controlled Win Rate. 
    We highlight the best performing model in each column in bold 
    and underscore the second best performing approach.
    Both \ours methods (\ours-$\pi_\text{ref}$ and \ours-$\pi_\text{old}$) are consistently better than the baselines except for one case. }
    \label{tab:if-results}
\end{table*}
\begin{table*}
    \centering
    \begin{tabular}{lrrr@{}cr@{}c}
    \toprule
     & \multicolumn{2}{c}{\textbf{Expert Rub.}} && \textbf{GPQA-D} && {\textbf{GSM8K}} \\ \cmidrule{2-3} \cmidrule{5-5} \cmidrule{7-7}
    \textbf{Model} & \multicolumn{1}{c}{Score} & \multicolumn{1}{c}{WR} && \multicolumn{1}{c}{Acc.} && Acc. \\
    \rowcolor{gray!20} \multicolumn{7}{c}{\textit{Baselines}} \\
    Qwen-2.5-7B-Instruct                & 33.6 & 31.9 && 34.7 && 79.2 \\
    ~~~~+ LLM-Judge Score              & 36.7 & 44.0 && 34.5 && 79.1 \\
    ~~~~+ Offline Rubrics (Synthetic)               & 37.1 & 46.4 && 36.6 && 79.2 \\
    ~~~~+ Offline Rubrics (Human-written)           & 39.2 & 51.8 && 36.2 && 79.9 \\
    ~~~~~~~~+ Universal Requirements    & 39.7 & 53.3 && 36.6 && \underline{80.1} \\
    ~~~~~~~~+ Pointwise Extraction      & 40.9 & \underline{57.1} && 33.6 && 78.3 \\ 
    \rowcolor{gray!20} \multicolumn{7}{c}{\textit{Our Methods}} \\
    ~~~~~~~~+ \ours-$\pi_\text{ref}$             & \underline{41.4} & \textbf{61.0} && \underline{37.6} && 80.0 \\
    ~~~~~~~~+ \ours-$\pi_\text{old}$                & \textbf{41.5} & 56.5 && \textbf{38.1} && \textbf{80.5} \\
    \bottomrule
    
    \end{tabular}
    \caption{Results on training on the Expert rubrics.     
    WR stands for win rate and Acc. stands for accuracy.
    We highlight the best performing model in each column in bold and underscore the second best performing approach.
    Both \ours methods outperform the baselines.}
    \label{tab:prof-results}
\end{table*}
\subsection{Results and Discussion}
\label{sec:results}

Figure~\ref{fig:training-evaluation-curves} shows the training curves for the Generalist and Expert datasets.
Training with rubrics consistently scores higher and is more sample efficient than using LLM-Judge scores.
More interestingly, 
adding the elicited rubrics during training (\ours) improves the performance of the model on the evaluation sets of both datasets, which only contain human-written rubrics.

Table~\ref{tab:if-results} and Table~\ref{tab:prof-results} present the results on a set of instruction-following and reasoning benchmarks, respectively.
Training with Offline Rubrics (Human) is improving the performance of the model on all the respective datasets 
with the only exception being length controlled win rate on AlpacaEval (28.2\% vs 26.9\%).
Importantly, training with Offline Rubrics (Human) is (a) always better than using LLM-Judge scores across all benchmarks, 
and (b) is better than using synthetic rubrics across 7 out of 9 evaluation metrics. More interestingly, 
adding the elicited rubrics to the offline rubrics (human-written) during training (\ours) further boosts performance across both instruction-following and reasoning benchmarks.
On AlpacaEval, for instance, \ours-$\pi_\text{ref}$ increases the win rate from 46.4\% to 55.0\%, while also improving the length-controlled win rate (LC-WR) from 28.0\% to 31.5\% reflecting better quality responses in general.

When compared against other baselines,
\ours is consistently better than Universal Requirements across all benchmarks.
This is interesting because it suggests that sample-grounded elicited rubrics are more effective than augmenting the rubrics with a set of fixed criteria that fail to capture the nuances of individual prompts and remain static as the policy evolves during training. While adding pointwise extracted rubrics also often improves over offline rubrics, it is still surpassed by \ours (48.1 vs. 54.0 and 55.0 on AlpacaEval, 51.1 vs. 55.7 and 56.5 on Arena-Hard).
\ours leverage pairwise differences to highlight discriminative properties that distinguish a better response from a worse one rather than relying on a single response.

\subsection{Qualitative Analysis}
\label{sec:qualitative-analysis}

We conduct a qualitative analysis of the elicited  criteria and contrasted it with human-written rubrics. 
In summarizing the differences, we apply an LLM-based comparison of rubric updates (between the initial rubrics and rubrics at the last epoch) followed by clustering 
to identify recurring themes. We observe several consistent types of improvements in elicited criteria emerge. First, elicited criteria frequently introduced 
\emph{evidence grounding (e.g., The response includes only categorically relevant, evidence-backed details.)}, \emph{reproducibility} \emph{(e.g., The response avoids any process that can't be reproduced without modern technology.)}, and \emph{holistic anti-gaming criteria} \emph{(e.g, The response avoids over-specification and over-enumeration.)}, 
broadening the evaluative focus beyond surface-level correctness. Second, many criteria emphasize 
\emph{practicality and real-world feasibility} rewarding implementation readiness and 
resource awareness. Third, we observe that the addition of meta-criteria such as \emph{structural 
organization}, \emph{causal reasoning}, and \emph{uncertainty handling} enhance the rubric's coverage of 
system-level and methodological dimensions.

Overall, the new criteria highlight that online elicitation tends to expand and strengthen rubrics 
over time. Instead of remaining fixed, criteria adapt dynamically as new errors or weaknesses are 
exposed, leading to more comprehensive and resilient evaluation standards. A complete list of clusters 
with proportions is presented in \cref{app:qualitative_clusters}.

\section{Conclusion}
\label{sec:conclusion}

We have described \ours, a framework for dynamically eliciting new criteria from pairwise comparisons of responses during reinforcement learning. Unlike static rubrics which may be incomplete or become obsolete as training progresses, our approach aims to continuously surface overlooked errors or emerging desired properties. This yields robust gains across expert and generalist domains. Our results show improvements of up to 8 percentage points over training exclusively with human-written rubrics on AlpacaEval, GPQA and Arena-Hard. By moving rubric elicitation online, OnlineRubrics adapts as training evolves, capturing emergent behaviors and strengthening alignment beyond what fixed rubrics allow.


\bibliography{onlinerubrics}
\bibliographystyle{abbrvnat}
\appendix
\crefalias{section}{appendix}
\crefalias{subsection}{appendix}  
\crefalias{subsubsection}{appendix}

\section{Proof for \cref{prop:gradient}}
\label{app:proof_gradient}
\begin{proof}
    \begin{align*}
        g_U - g_{R_t} &= \mathbb{E}_{(x,o)} \Big[\nabla_\theta \log\pi_\theta(o|x) \big(U-R_t\big)\Big] \\
        &= \mathbb{E}_{(x,o)} \Big[\nabla_\theta \log\pi_\theta(o|x) \big(Y - \mathbb{E}_{(x,o)}\big[Y\big]\Big] \qquad \text{where } Y = U - R_t 
    \end{align*}
because $\mathbb{E}_{(x,o)}\big[\nabla_\theta \log\pi_\theta(o|x)\big] = 0$ we can center $Y$ without changing the expectation. Then
\begin{align*}
        \Big\lVert g_U - g_{R_t} \Big\rVert_2 &= \Big\lVert \mathbb{E}_{(x,o)} \Big[\nabla_\theta \log\pi_\theta(o|x) \big(Y - \mathbb{E}_{(x,o)}\big[Y\big]\big)\Big]\Big\rVert_2 \\
        &\leq \sqrt{\mathbb{E}\Big[\big\lVert \nabla_\theta\log_{\pi_\theta} \big\rVert^2\Big]}\sqrt{Var(Y)} \qquad \text{by Cauchy-Schwarz} \\
        &= \sqrt{\mathbb{E}\Big[\big\lVert \nabla_\theta\log_{\pi_\theta} \big\rVert^2\Big]}\sqrt{Var(U - R_t)} \\
        &= \sqrt{\mathbb{E}\Big[\big\lVert \nabla_\theta\log_{\pi_\theta} \big\rVert^2\Big]}\sqrt{\mathbb{E}\big[(U - R_t)^2\big]} \\
        &= \sqrt{\mathbb{E}\Big[\big\lVert \nabla_\theta\log_{\pi_\theta} \big\rVert^2\Big]}\big\lVert w_I\big\rVert_1
\end{align*}

\end{proof}

\section{Data Samples}
\label{app:data_samples}

\begin{figure}[h]
    \centering
    \includegraphics[width=0.9\textwidth]{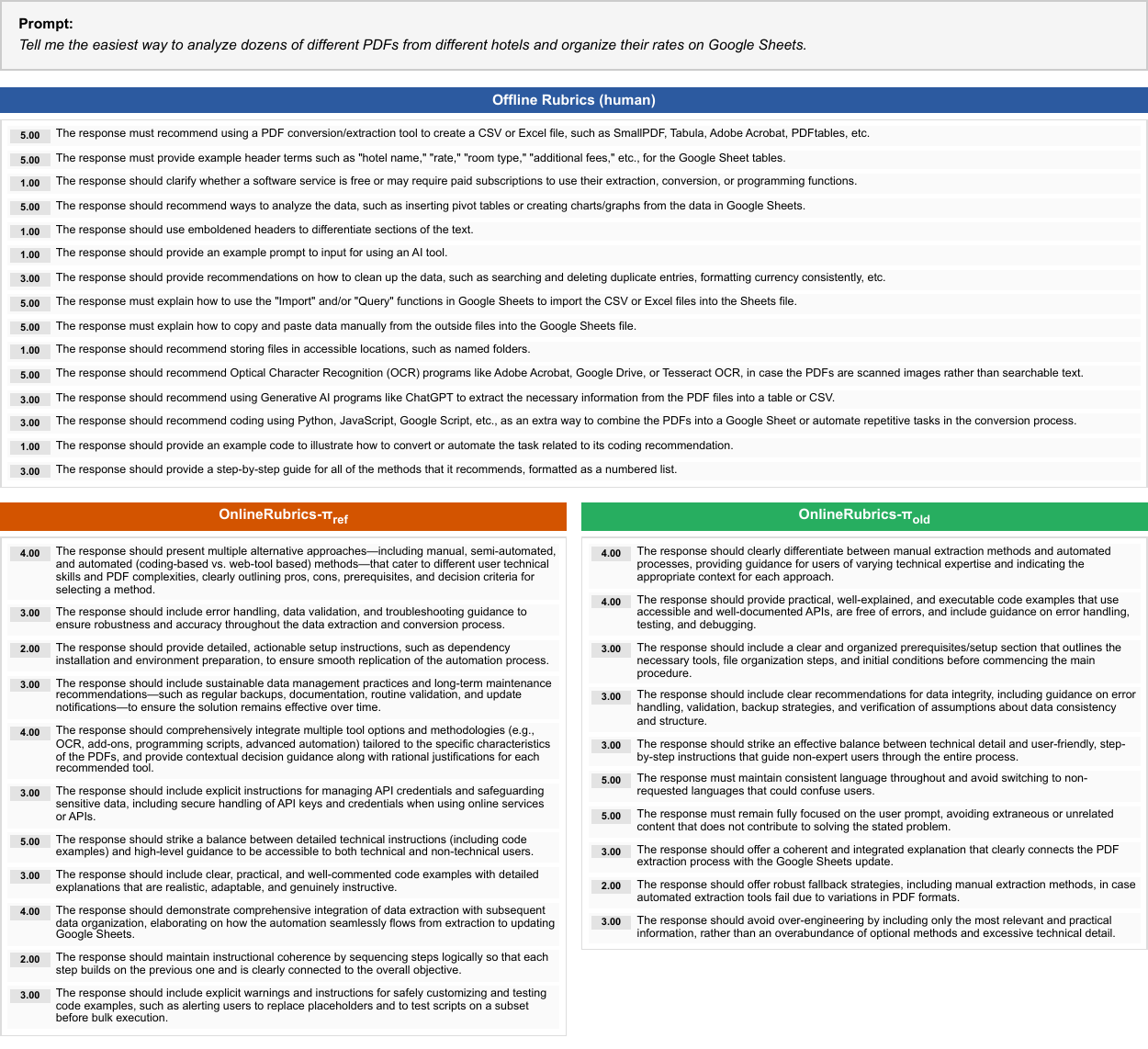}
    \caption{Data sample from the Generalist Rubrics dataset.}
    \label{fig:data_samples}
\end{figure}

\begin{figure}[h]
    \centering
    \includegraphics[width=0.9\textwidth]{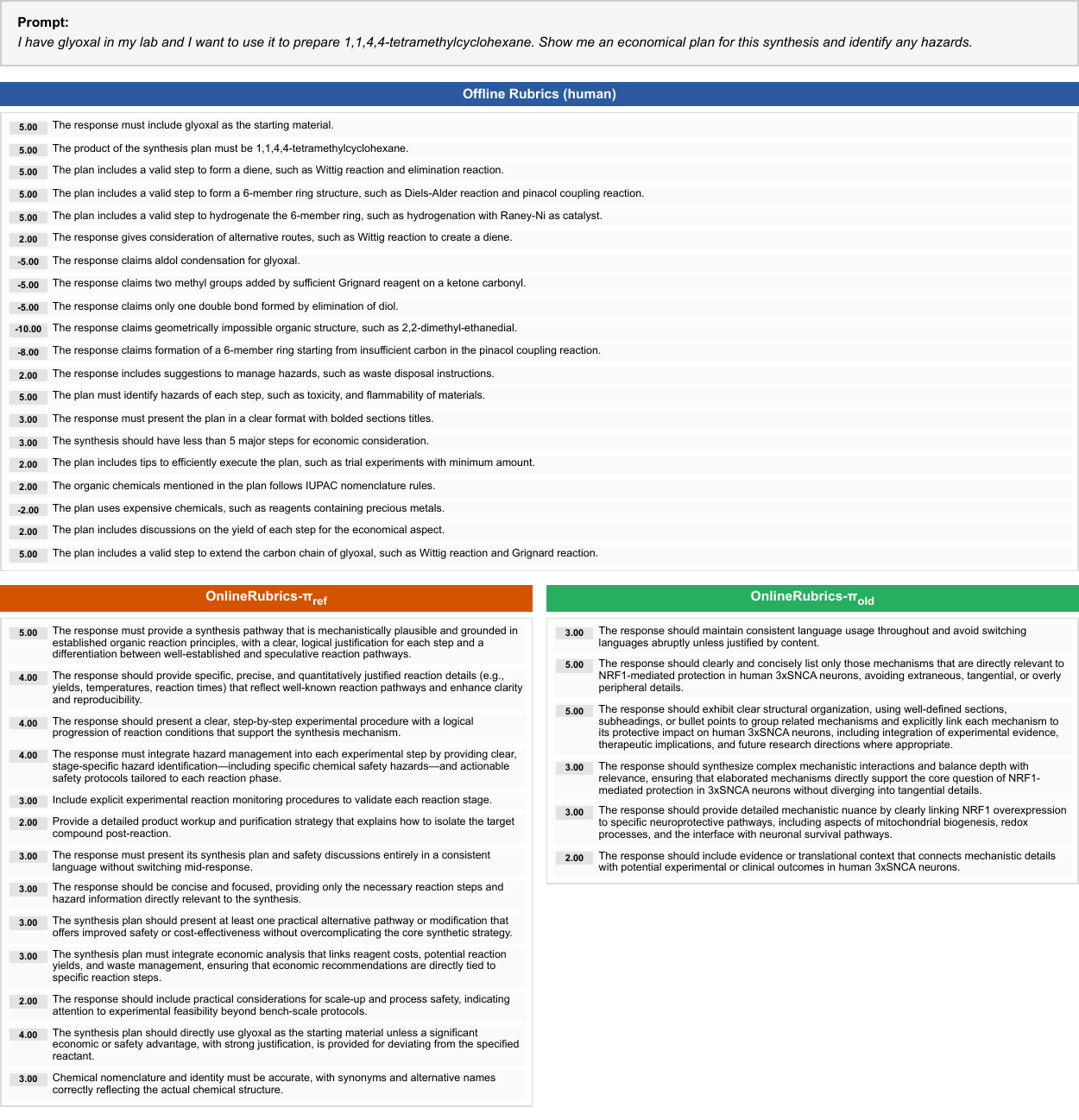}
    \caption{Data sample from the Expert Rubrics dataset.}
    \label{fig:data_samples_2}
\end{figure}

We provide two samples showing sampled rollouts from current and reference policies, along with human and elicited rubrics in \cref{fig:data_samples,fig:data_samples_2}. Each criteria are preceded with its importance weight which range between 1-5 for Generalist and -10 and 10 for Expert sets.

\section{Experimental Settings}
\label{sec:experimental_settings}

\subsection{Training Settings.}
We train Qwen-2.5-7B-Instruct~\citep{qwen2025qwen25technicalreport} on the training set of the Generalist and Expert Rubrics datasets for three epochs. 
Training follows the GRPO procedure described in \cref{sec:background}, with 16 rollouts generated per sample. 
We use GPT-4.1-mini as the \llmgrader and o3-mini as the \llmextractor, performing eight pairwise comparisons per instance. 
Optimization uses a learning rate of $5e-6$ with a warmup ratio of 0.1. 
KL-divergence regularization is applied with a coefficient of 0.01. 
All experiments are conducted on 8 NVIDIA H100 GPUs with per-device batch size of 6 and gradient accumulation of 2 steps (effective batch size is 96).

\subsection{Evaluation Settings.}

\paragraph{Generalist and Expert Rubrics Datasets.}
We calculate the score and win rate (vs. the reference policy) on the evaluation set of the Generalist and Expert Rubrics datasets.
Score is calculated as explained in \cref{eq:rubric_reward_reduction}. We use GPT-4.1-mini as the \llmgrader.
We use Gemini-2.5-Pro as the LLM-Judge that picks the winner between the two responses.
For each sample, we run the judge twice by flipping the order of the two responses.
If the judge picks the same response twice, we consider it as a win. 
The prompt for the judge is given in \cref{app:system_prompts}. 

\paragraph{AlpacaEval.}
We use the evaluation script\footnote{\url{https://github.com/tatsu-lab/alpaca_eval}} from \citep{alpaca_eval} to calculate the win rate
and length controlled win rate on the evaluation set of AlpacaEval using the default settings.

\paragraph{Arena-Hard.}
We use the evaluation script\footnote{\url{https://github.com/lmarena/arena-hard-auto}} from \citep{li2024crowdsourced, arenahard2024} to calculate the win rate (vs. the reference policy) on the evaluation set of Arena-Hard.
We use GPT-4.1 as the LLM-Judge.

\paragraph{GPQA-Diamond.}
We use simple-evals\footnote{\url{https://github.com/openai/simple-evals}} for evaluation on GPQA-Diamond~\citep{rein2024gpqa} and report the average accuracy across 4 runs.

\paragraph{GSM8K.}
We use lm-evaluation harness~\cite{eval-harness} to calculate the strict match accuracy on the evaluation set of GSM8K~\citep{gsm8k}.

\section{System Prompt Templates}
\label{app:system_prompts}

\begin{figure}[h]
    \centering
    \includegraphics[width=0.97\textwidth]{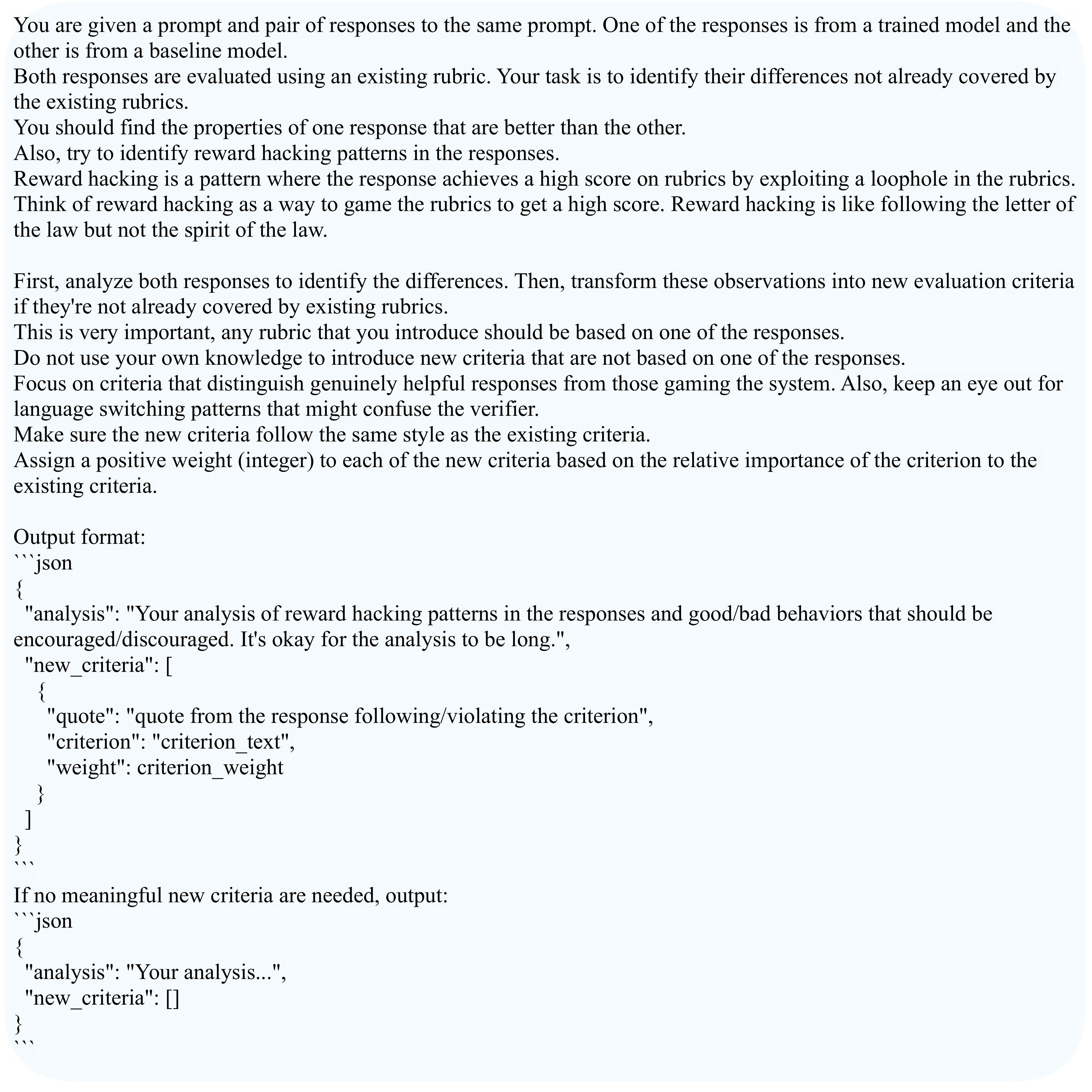}
    \caption{Full system prompt template used for \llmextractor.}
    \label{fig:extractor_prompt_full}
\end{figure}

\begin{figure}[h]
    \centering
    \includegraphics[width=0.97\textwidth]{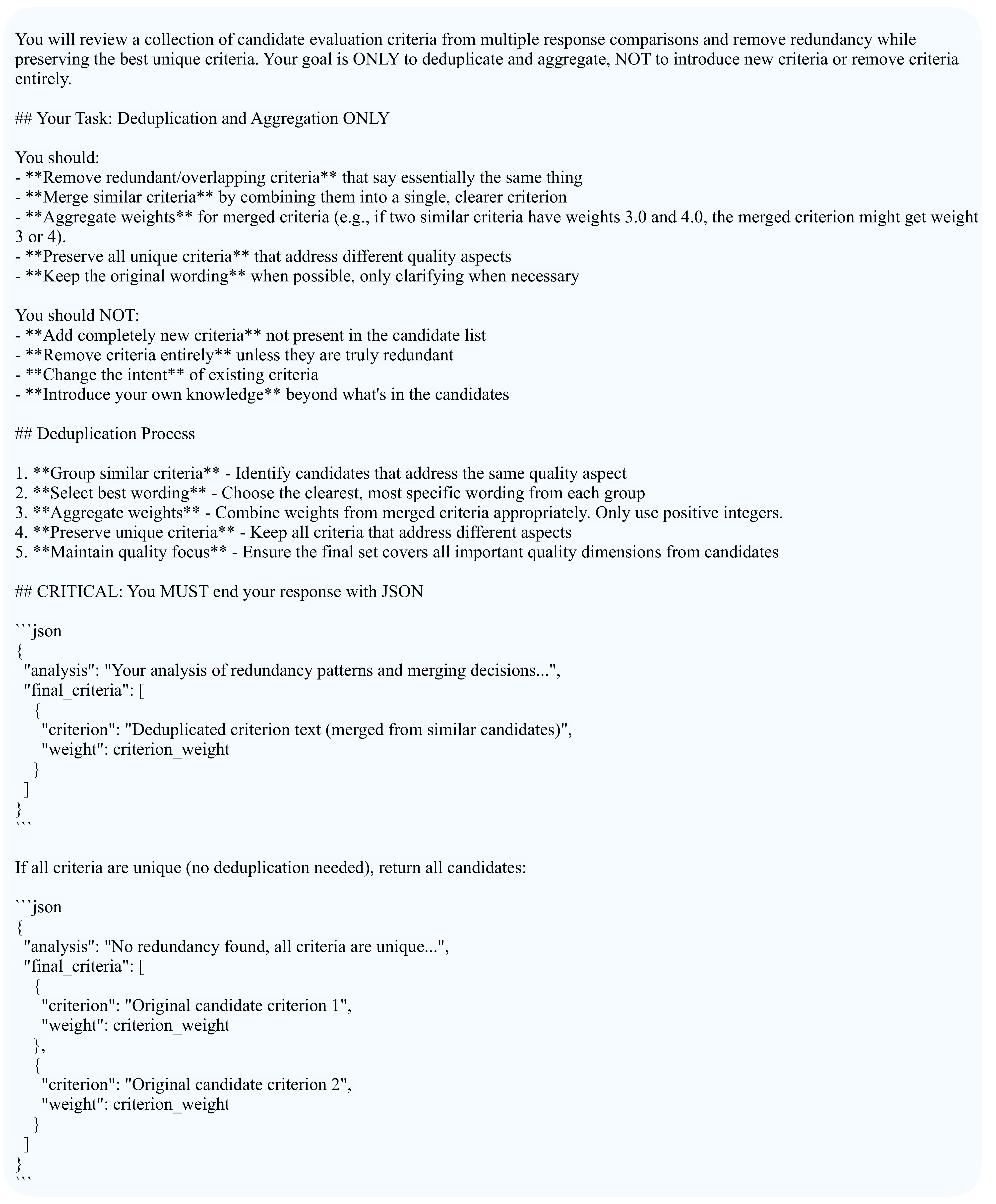}
    \caption{Full system prompt template used for de-duplicating extracted criteria.}
    \label{fig:deduplication_prompt}
\end{figure}

\begin{figure}[h]
    \centering
    \includegraphics[width=0.97\textwidth]{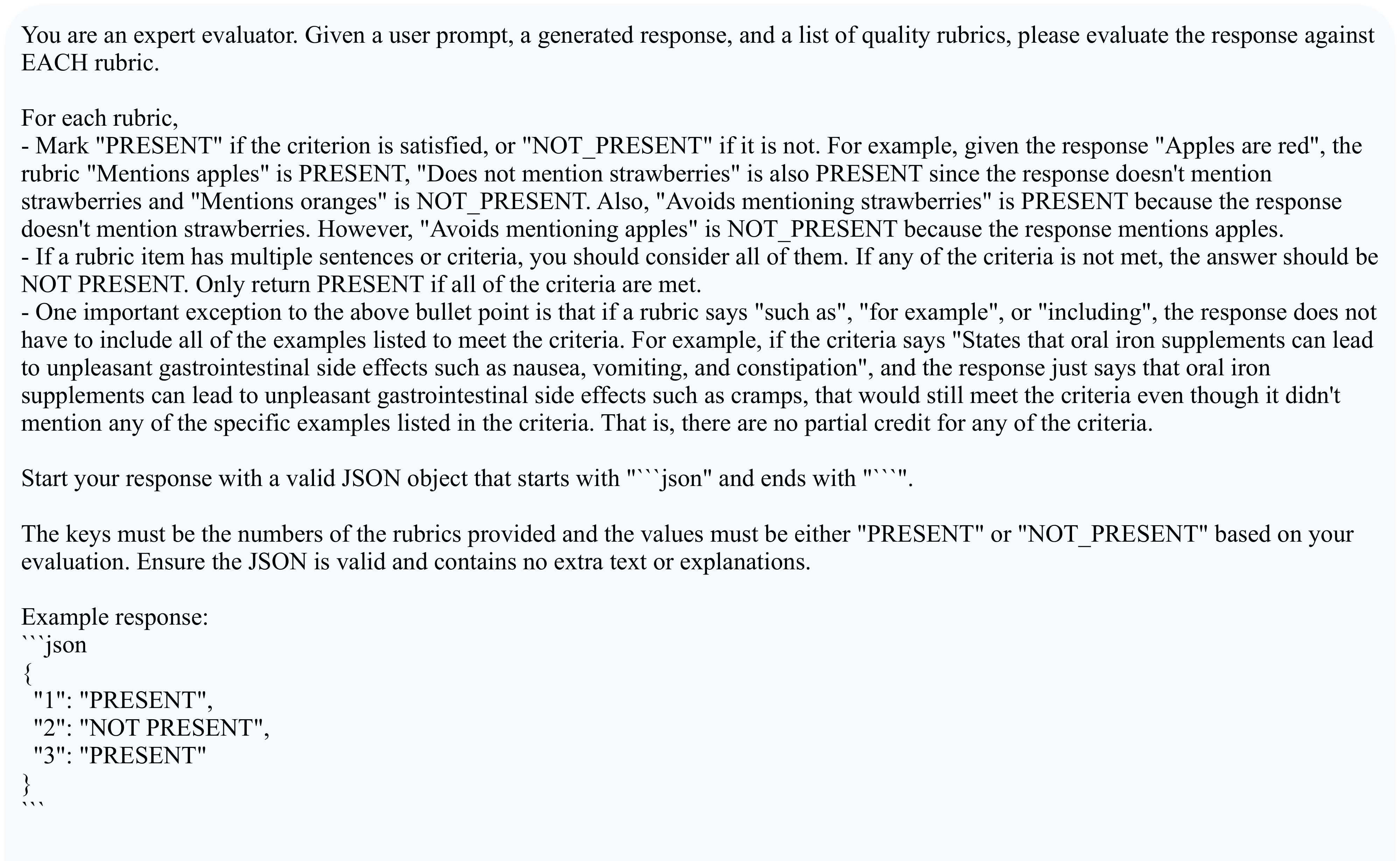}
    \caption{Full system prompt template used for \llmgrader.}
    \label{fig:grader_prompt}
\end{figure}

Figures~\ref{fig:extractor_prompt_full} and \ref{fig:deduplication_prompt} show the system prompt templates used for \llmextractor and de-duplicating extracted criteria, respectively.
We use the system prompt provided in \cref{fig:grader_prompt} for \llmgrader.

\begin{figure}[h]
    \centering
    \includegraphics[width=0.97\textwidth]{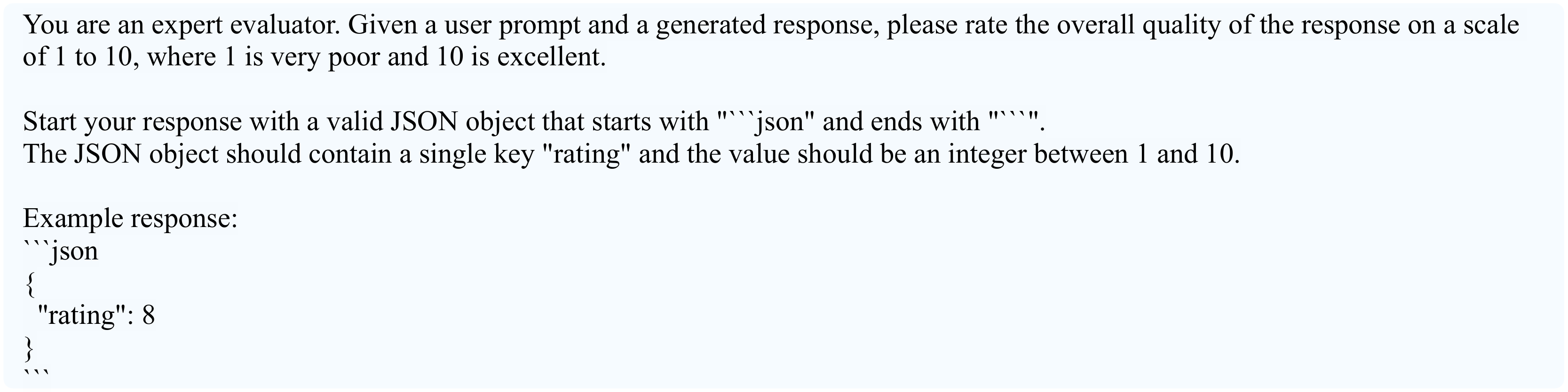}
    \caption{Full system prompt template used for LLM-Judge Score.}
    \label{fig:llm-score-prompt}
\end{figure}

\begin{figure}[h]
    \centering
    \includegraphics[width=0.97\textwidth]{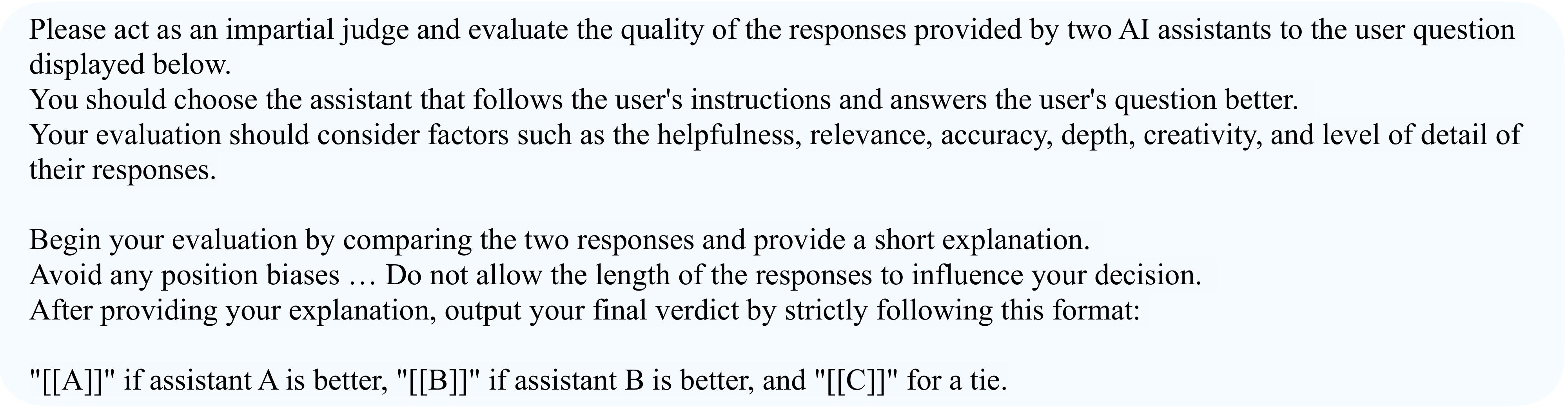}
    \caption{Full system prompt template used for LLM-Judge for win rates.}
    \label{fig:llm-judge-prompt}
\end{figure}

\begin{figure}[h]
    \centering
    \includegraphics[width=0.97\textwidth]{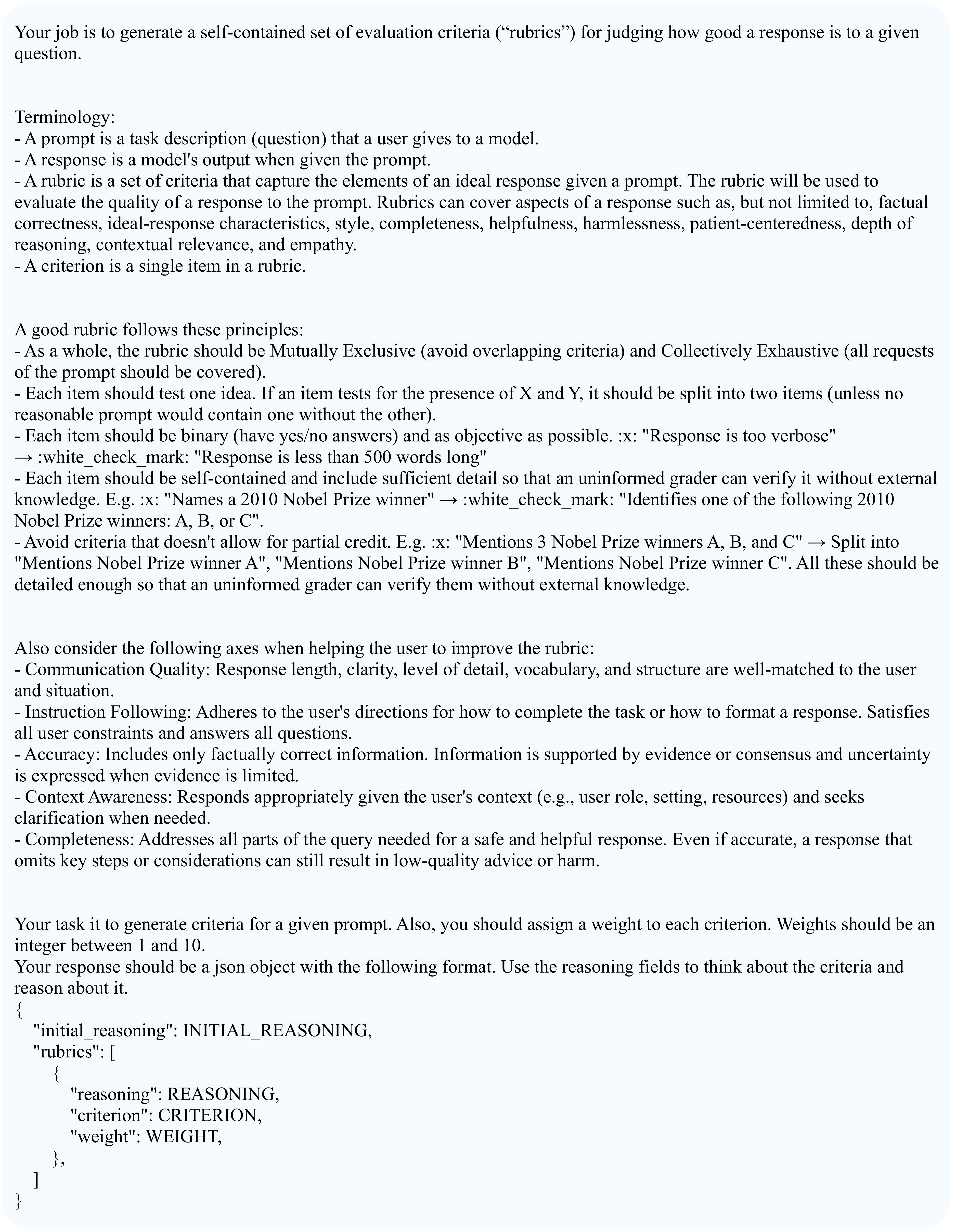}
    \caption{Full system prompt template used to generate synthetic rubrics.}
    \label{fig:synth-gen-prompt}
\end{figure}

Figures~\ref{fig:llm-score-prompt} and \ref{fig:llm-judge-prompt} show the system prompt templates used for LLM-Judge Score and LLM-Judge for win rates, respectively.
We use the system prompt provided in \cref{fig:synth-gen-prompt} to generate synthetic offline rubrics.

\section{Qualitative Rubric Clusters}
\label{app:qualitative_clusters}

\begin{figure}
    \centering
    \includegraphics[width=0.97\textwidth]{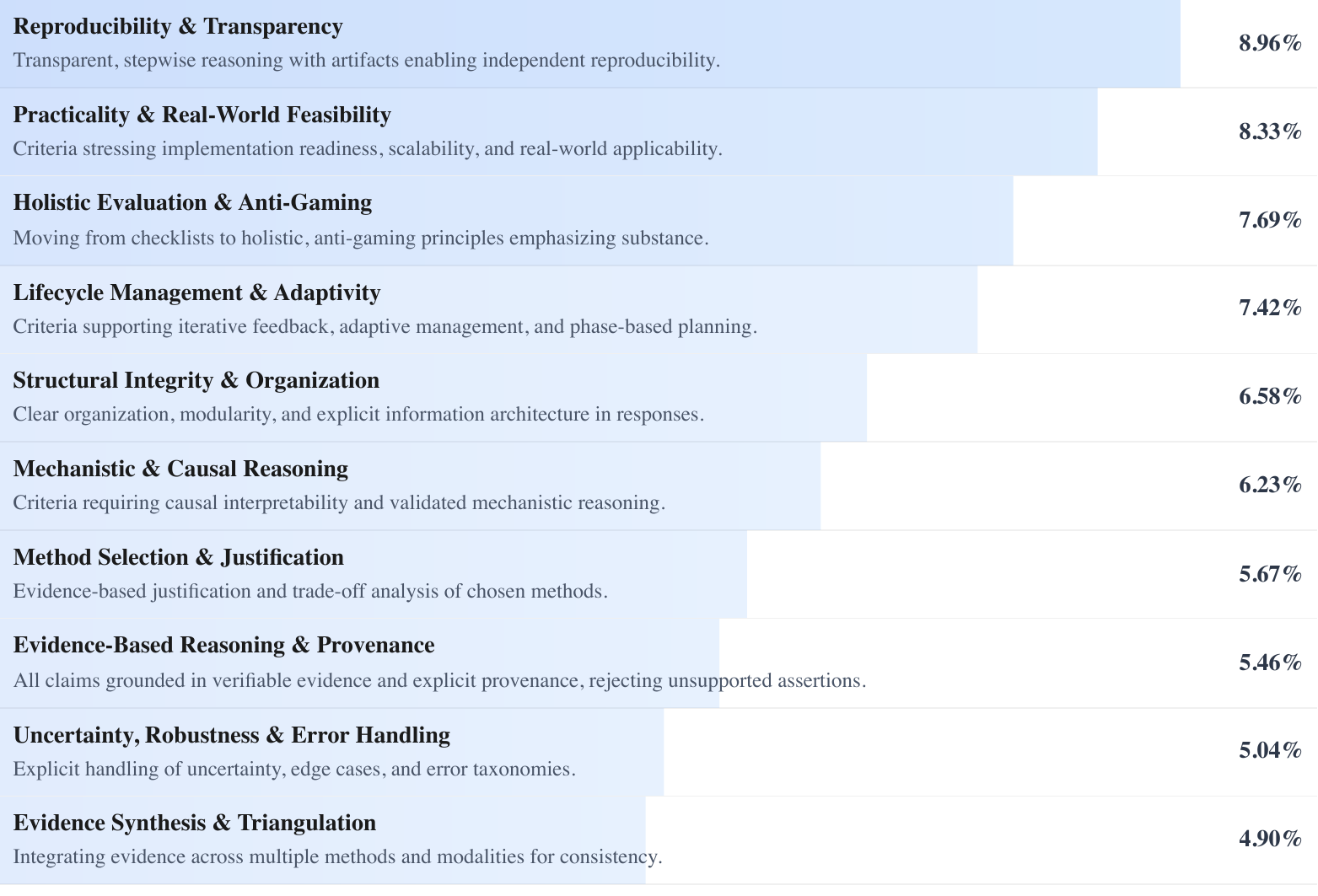}
    \caption{Top-10 most frequent clusters of rubric criteria elicited via \ours. Each cluster is shown with a short description and its share of samples, sorted by proportion.}
    \label{fig:qualitative_clusters}
\end{figure}

We report the clusters of rubric refinements observed during online elicitation. 
Figure~\ref{fig:qualitative_clusters} lists each cluster with its name, a concise description, and its share of samples, 
sorted by proportion.

\end{document}